\documentclass{article}

\usepackage{PRIMEarxiv}

\usepackage[utf8]{inputenc} 
\usepackage[T1]{fontenc}    
\usepackage{hyperref}       
\usepackage{url}            
\usepackage{booktabs}       
\usepackage{amsfonts}       
\usepackage{nicefrac}       
\usepackage{microtype}      
\usepackage{lipsum}
\usepackage{fancyhdr}       
\usepackage{graphicx}
\graphicspath{{media/}}     
\usepackage{url,hyperref,lineno,microtype,subcaption}
\usepackage[onehalfspacing]{setspace}

\title{A Gamified Interaction with a Humanoid Robot to explain Therapeutic Procedures in Pediatric Asthma
\thanks{\textit{\underline{Citation}}: 
\textbf{Laura Montalbano, Agnese Augello,  Giovanni Pilato, Stefania La Grutta, arXiv preprint}} 
}

\author{
  Laura Montalbano, Agnese Augello *,  Giovanni Pilato *, Stefania La Grutta **\\
  Institute for High Performance Computing and Networking (ICAR) *\\
Institute of Traslational Pharmacology (IFT) **\\
  CNR \\
  Palermo, Italy\\
  \texttt{dr.montalbanolaura@gmail.com, agnese.augello@icar.cnr.it} \\
  \texttt{giovanni.pilato@icar.cnr.it, stefania.lagrutta@ift.cnr.it}   
}

\begin{document}
\maketitle
\begin{abstract}

In chronic diseases, obtaining a correct diagnosis and providing the most appropriate treatments often is not enough to guarantee an improvement of the clinical condition of a patient. Poor adherence to medical prescriptions constitutes one of the main causes preventing achievement of therapeutic goals. This is generally true especially for certain diseases and specific target patients, such as children. An engaging and entertaining technology can be exploited in support of clinical practices to achieve better health outcomes. Our assumption is that a gamified session with a humanoid robot, compared to the usual methodologies for therapeutic education, can be more incisive in learning the correct inhalation procedure in children affected by asthma. In this perspective, we describe  an interactive module implemented on the Pepper robotic platform and the setting of a study that was planned in 2020 to be held at the  Pneumoallergology Pediatric clinic of CNR  in Palermo. The study was  canceled due to the COVID-19 pandemic.   Our long-term goal is to assess, by means of a qualitative-quantitative survey plan, the impact of such an educational action, evaluating possible improvement in the adherence to the treatment. 
\end{abstract}

\section{Introduction}
In chronic diseases requiring long-term treatments, therapeutic adherence is an essential means towards symptom control  \cite{sabate2003adherence}. When it is children who are affected by a chronic disease, managing this aspect becomes more challenging. It is indeed necessary to educate not only families but also, most importantly, children about how to correctly self-administer their daily therapy and what to do in case of emergency, considering their frequent absence from home due to attending school, playing with friends, and other activities. Asthma is one of the most widespread diseases in pediatric age \cite{ferrante2018burden}.
Being a multifactorial disease especially in presence of comorbidity \cite{cibella2015burden}, it has a significant influence on both the patients’ and their families’ quality of life. Inhalation reliever and controller therapy represents the main treatment for asthma, and inhalation devices offer advantages such as ease of transport, rapid and targeted action \cite{albano2013th17} and minimal side effects compared to a systemic drug \cite{2}. However, it is quite difficult to execute a correct inhalation technique \cite{3}. It has been observed, in this respect, that an incorrect inhalation technique is related to uncontrolled symptoms of asthma \cite{4}. In addition to this, studies have found that up to 85\% of healthcare professionals do not seem to be capable of educating patients in the correct inhalation technique or showing them how to execute it. \cite{5}. GINA 2017 International Guidelines for asthma management recommend self-management programs that improve outcomes in patients with asthma. To date, though, these programs have not yet been sufficiently implemented into clinical practice or entirely accepted by patients. In this respect, the rapid technological evolution that has occurred during the last few decades offers new opportunities to design and provide self-management initiatives \cite{6}\cite{7}. In relation to this, the late ‘90s witnessed the birth and development of  so-called “electronic health management”, known as e-health \cite{8}, encompassing a vast area regarding medical IT and public health, aimed at providing and/or improving healthcare services and information through the internet and its related technologies \cite{9}. \\
The goal of e-health is to improve healthcare quality, quality of life and therapeutic adherence in chronic diseases \cite{8} \cite{montalbano2019targeting} \cite{licari2019impact} and includes mobile-health, telehealth and AI \cite{10}.In the past few decades we have been facing and interacting with more and more technology. Technology is evolving very fast and has been changing our lives significantly compared to the past. Robotics is an emerging field, especially in healthcare and therapy due to its potentiality and capacity of engagement for different roles. Robots can be therapeutic and act as companions, becoming an extension of the therapist  \cite{11}.
Nowadays they are increasingly used in healthcare, where they are known as Socially Assistive Robots (SARs). They are used to support people in correctly performing the assigned therapy \cite{ferrante2021social} \cite{winkle2018social}, acting as motivators, providing feedbacks \cite{rodriguez2018emotional} and, in some cases, also offering  a physical support. A study conducted by Winkle et al. \cite{winkle2019mutual} focused on the importance of a participatory design of  human-robot interaction and the ‘mutual shaping’ process of the stakeholders involved in therapy programs, such as carers, healthcare professionals, family and friends. The main results of the study were guidelines on the adaptation of the therapy to make best use of robots and the identification of social and contextual factors likely to affect/be affected by use of an SAR. Robots are particularly effective with children, since they are more attracted by this kind of technology, in approaching therapy  \cite{martelo2017social} \cite{yap2017reach}. Moreover, a robot can be a way to distract children to reduce their suffering and distress during healthcare procedures  \cite{beran2013reducing}. 
\\
Focusing on the specific scenario of asthma, there are several studies proving that a gamified approach, with the support of innovative technologies, can support and motivate children in doing the therapy. Original apps and games have been proposed with different purposes: to analyze the cough, to provide information or to make the child more autonomous in the therapy. To the best of our knowledge there are no previous studies conducted in a Pediatric Pneumoallergology clinic on the effectiveness of the interaction with a humanoid robot to improve adherence to treatment in children affected by asthma. We found a similar approach in the robotic companion named JOE developed by the Ludocare company \cite{brage2018joe}. Although JOE is not exactly a robot, it is considered one by children interacting with it. JOE alerts when it’s therapy time, it guides the child in the gestures to perform with his or her medications and motivates him or her . JOE explains how the treatment is to be treated, rhythmizes the treatment with sound and by blinking its eyes, and collects feedbacks from the child, asking if he or she was able to get treatment. If the child responds positively, it offers a reward, which can consist in a joke, a video, a story or something else. The company will statistically analyze the interactions but at the moment there are no results and the setting is different, since in our proposal the interaction happens in the clinic at first diagnosis. Another similar approach is the case of a Pepper robot located at a pharmacy to explain with some videos / a video how to use inhalators in patient affected by asthma  \cite{ishiguro2018development}. In this case too, after an explanatory phase the robot runs a quiz to test if the procedure has been understood. Again, the setting and the target is different and there is no study proving the effectiveness of the proposed approach. The primary aim of the study presented in this paper is to assess improvements in outpatient children with asthma with respect to the inhalation technique, through the use of robotics. Its secondary objective is to assess children’s ability to correctly self-administer their therapy, resulting in improved therapeutic adherence compared to children who receive the usual care. This study would provide an innovative approach to therapeutic education in pediatric asthma, and help the clinician provide information regarding correct management of the disease.

\section{Study Design}

This is a pilot study to assess the intervention of a humanoid robot in instructing children with asthma on the correct execution of the inhalation technique. Male and female patients aged between 6 and 11 diagnosed with asthma will be enrolled consecutively by Clinical and Environmental Epidemiology of Pulmonary and Allergic Disease (CEEPAPD) at Institute for Research and Biomedical Innovation (IRIB), National Research Council (CNR) in Palermo. Patients will be randomly assigned either to a group that will involve the intervention of the humanoid robot (Group 1) or a group that will receive the usual care (Group 2). During their first  check, after being diagnosed with asthma through a clinical and instrumental assessment, all participants will be invited to fill out standardized disease-specific questionnaires aimed at evaluating their symptom control as well as their own and their families’ quality of life. The robot will thoroughly explain the correct execution of the therapy to Group 1, whereas a clinician will provide the same detailed information to Group 2. Following the intervention, instructions will be verified through a tablet game for Group 1 and paper and pencil support for Group 2; in addition to this, children will be asked to repeat the inhalation technique previously observed without receiving any support from the clinician. Patients enrolled in this study will not be affected by any specific learning disorders or executive dysfunction. The clinician will verify that the patient correctly carries out each one of the explained steps and will take note of any possible mistakes made during execution of the inhalation technique. A follow-up will be performed after 3 months, in which the learning of the correct inhalation technique will be verified and disease-specific questionnaires will be administered to evaluate possible improvement of the disease outcomes. The people involved in the study, present in the clinic, will be a doctor, a pediatric psychologist and an engineer.

\section{Gamified interaction with the Pepper robot}

In what follows, we describe an interactive module implemented on a Pepper robot to explain the inhalation procedure. We use Aldebaran’s Pepper robot as the robotic platform. Pepper is equipped with a tablet that is particularly suitable for this kind of activity. The module has been designed to catch children’s attention and describe the main steps of the procedure in an amusing way. The verbal description of the steps is accompanied by music and the visualization of appropriate images in the Pepper tablet. The description is also enriched with suitable gestures and expressiveness, to obtain a better engagement and at the same time boost the learning process. A chatbot-based dialogue component is used by Pepper to manage  interaction with children. We designed two activities. The first one consists in the explanation of the procedure through a succession of scenes, where each scene represents the description of one single step of the procedure. To avoid an information overload and at the same time simplify the explanation, we have summarized the procedure in four steps, also defining a simple acronym to help the child to memorize them. The interaction is in the Italian language, since the experimentation will be conducted in Italy. The chosen acronym is  ``ASMA'', corresponding to the first letters of the following sentences which constitute the four main steps of the procedure:

\noindent \textit{\textbf{A}gita la bomboletta e inseriscila nel distanziatore\\
\textbf{S}offia fuori l'aria per svuotare i polmoni \\
 \textbf{M}etti le labbra ben strette intorno al boccaglio e premi la bomboletta \\
\textbf{A}desso fai 5 respiri profondi dopo ogni puff e aspetta un po per rifare il puff }\\
\\
In what follows the Italian sentences are translated in English: 

\noindent \textit{Shake the canister and insert it into the spacer  \\
Breathe out the air to empty your lungs \\
Put your lips tightly around the mouthpiece and press the canister \\
Now take 5 deep breaths after each puff and wait a while to redo the puff}. \\


The second activity consists in a game. In the game the four steps are shown in the tablet in the wrong order. On the left side of the tablet the first four numbers are shown, while the right side depicts the images corresponding to the four possible steps. The child has to assign the correct figure to each of the four numbers on the left. The child interacts with Pepper by talking and touching the tablet in order to sort the steps correctly. For each step, the robot gives feedback to the child, inviting him or her to try again if he or she has made a mistake. The game ends when all the steps have been associated to a number by the patient. Pepper gives further feedback to the child showing the score obtained in the game.

	\begin{figure}[ht]
		\begin{center}
			\includegraphics[width=0.80\textwidth]{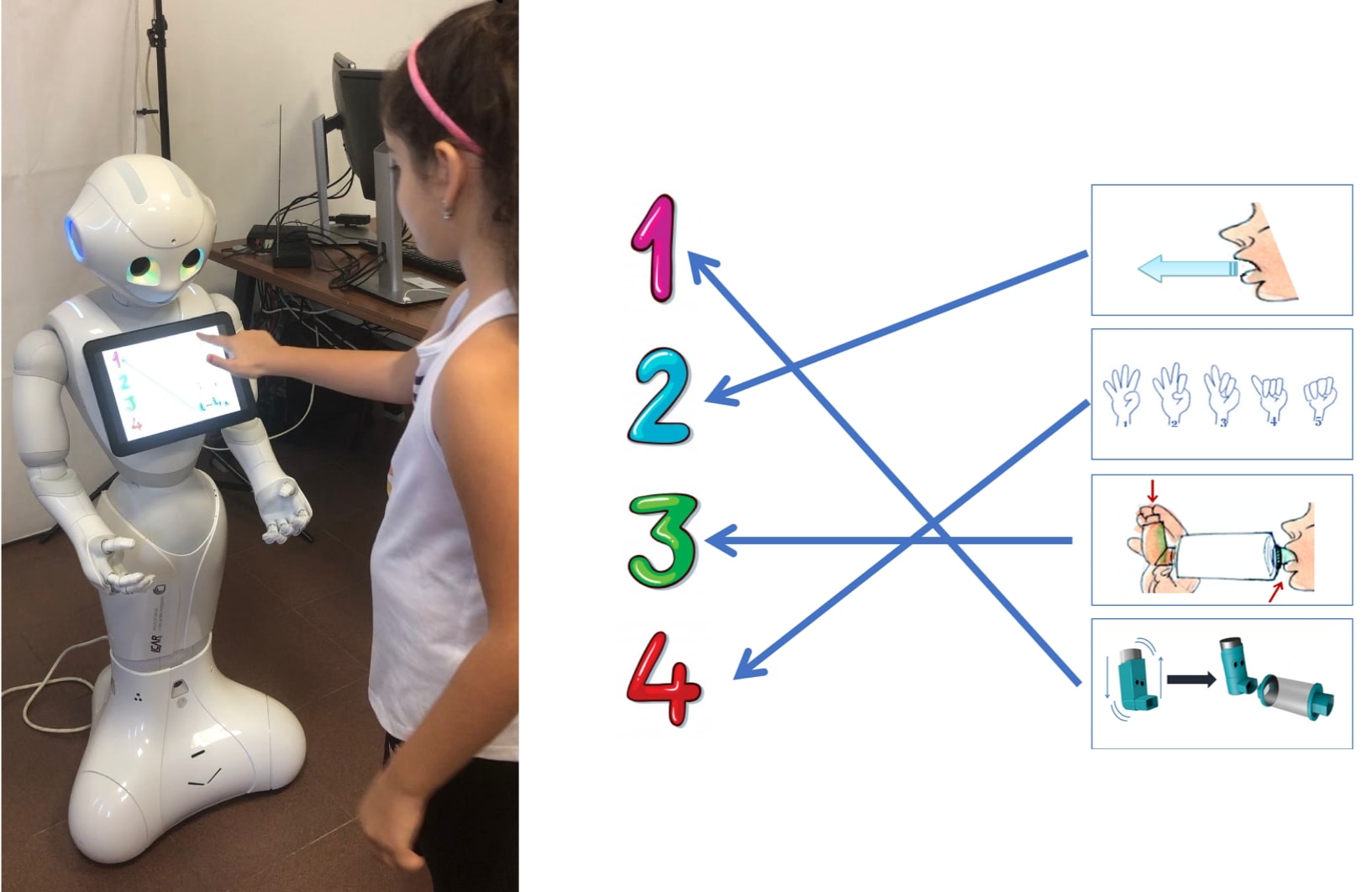} 
			\caption{A patient playing with Pepper and the detail of the tablet, showing the sequence of steps correctly sorted at the end of the game}
			\label{SyncMovment}
		\end{center}    
	\end{figure}

 
\section{Conclusion}

Education for correct execution of the inhalation technique should be an integral part of the transfer  of information useful for the management of the disease according to international asthma guidelines. However, this does not always occur: there are many barriers in outpatient settings, first of all limited time, followed by poor staff training, inadequate knowledge and above all the lack of demonstration material. As a result, health workers can investigate, through interviews and questionnaires, whether the child/parent has correct knowledge of the technique to determine if they need education in this regard. Self-assessment, however, almost always leads to imprecisions since children and parents often overestimate their abilities in performing the inhalation technique. Since this overestimation of their abilities can lead to poor self-management of the disease, causing failure to control symptoms, it is essential to instruct the child/parent in the inhalation technique and to evaluate its correct execution as accurately as possible \cite{volerman2019does}.
In this sense, new technologies could represent a valid support to health workers in patient/family education.

	\begin{figure}[ht]
		\begin{center}
			\includegraphics[width=0.85\textwidth]{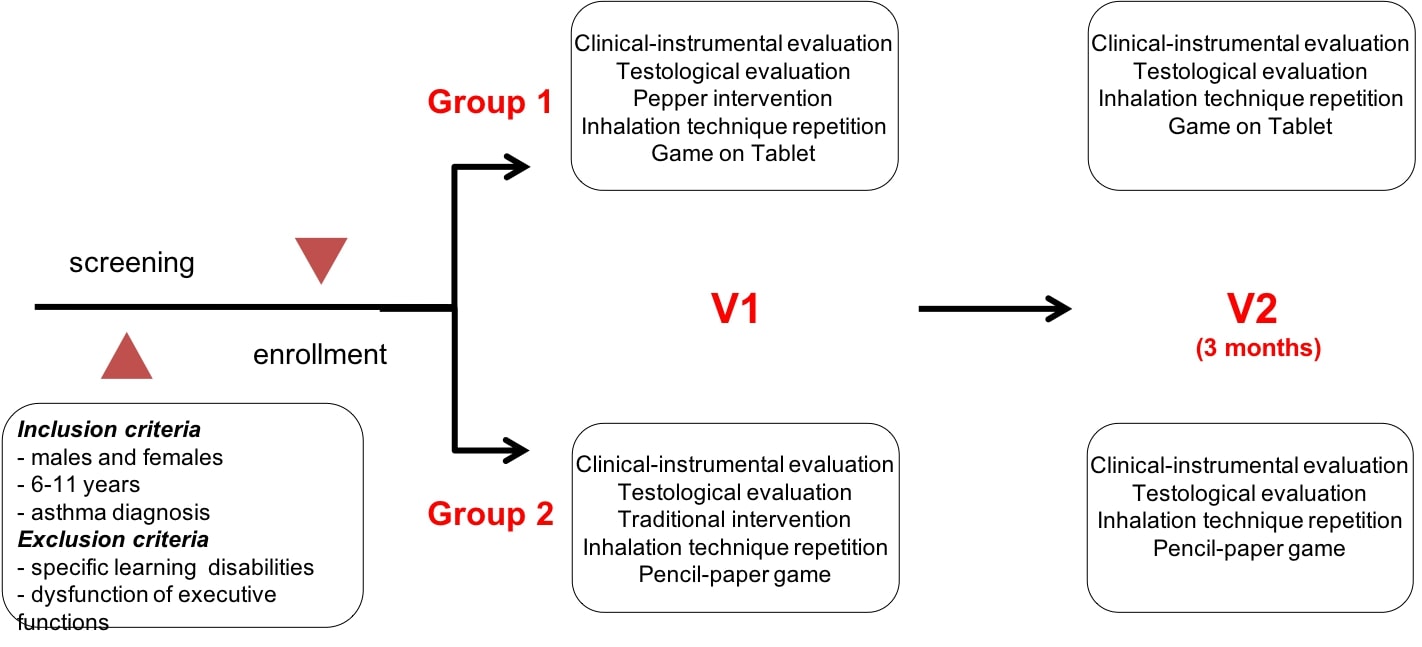} 
			\caption{The study design.}
			\label{images}
		\end{center}    
	\end{figure}

\bibliographystyle{unsrt}  
\bibliography{references}

\end{document}